\documentclass[letterpaper, 10pt, conference]{ieeeconf}  
\IEEEoverridecommandlockouts                              

\usepackage{graphics} 
\usepackage{epsfig} 
\usepackage{mathptmx} 
\usepackage{times} 
\usepackage{amsmath} 
\usepackage{amssymb}  
\usepackage{graphicx}
\usepackage{bm}
\usepackage{eucal}

\usepackage{float}
\usepackage[ruled,vlined]{algorithm2e}
\usepackage{pgfplotstable}
\usepackage{pgfplots}
\usepackage{tikz}
\usetikzlibrary[pgfplots.groupplots]
\usepgfplotslibrary{fillbetween}
\pgfplotsset{compat=1.5}
\usepackage{subcaption}
\usepackage{caption}
\usepackage{xcolor}
\usepackage{enumerate}
\usepackage{makecell}
\usepackage{colortbl}

\usepackage{graphicx}
\usepackage{tabularx,ragged2e,booktabs}
\usepackage{url}
\usepackage{siunitx}
\usepackage{rotating}
\usepackage{amssymb}
\usepackage{tabularx}
\usepackage{pgfplots}
\usepackage{csvsimple}
\usepackage{multirow}
\usepackage{cite}
\usepackage{float}
\usepackage{graphicx}
\usepackage{booktabs}
\usepackage{todonotes}
\usepackage{multirow}
\usepackage{amsmath}
\usepackage{wrapfig}
\usepackage{cancel}
\usepackage{titlesec}
\usepackage[font=footnotesize]{caption}
\SetKwComment{Comment}{$\triangleright$\ }{}

\SetCommentSty{mycommfont}

\makeatletter
\renewcommand\subsubsection{\@startsection{subsubsection}{3}{\z@}%
                                     {-3.25ex\@plus -1ex \@minus -.2ex}%
                                     {-1.5ex \@plus -.2ex}%
                                     {\normalfont\normalsize\itshape}}
\makeatother

\titlespacing*{\section}{0pt}{3pt}{1pt}
\titlespacing*{\subsection}{0pt}{1pt}{1pt}

\begin{document}
\title{Interactive Perception of Granular Materials via \\Robotic Scooping and Force–Torque Sensing}

\author {Samuli Hynninen$^1$, Mohammad Amman$^1$, Meri Mäkelä$^1$, Anirvan Dutta$^1$, Ville Kyrki$^1$
\thanks{$^1$Intelligent Robotics Group at the Department of Electrical Engineering and Automation, School of Electrical Engineering, Aalto University, Finland. {\tt\footnotesize\{firstname.lastname\}{@}aalto.fi}}}
\maketitle

\begin{abstract}
Granular materials exhibit complex and highly variable interaction dynamics, making their reliable manipulation challenging without knowledge of their physical properties. We present an interactive perception framework that infers granular material properties directly from force--torque measurements acquired during scooping. The proposed \textit{Latent Filter++} combines a bio-inspired frequency--temporal filter-bank representation with a causal latent dynamical model, allowing the representation to evolve as interaction observations become available while avoiding hand-crafted signal preprocessing. The learned latent representation captures physically meaningful variation in granular materials and supports estimation of properties including grain size and bulk density. Evaluated on $26$ diverse granular materials, the approach generalizes to materials excluded from training and enables few-shot adaptation of property-specific mappings using a small number of additional interaction samples. Finally, we demonstrate the relevance of the inferred properties to downstream manipulation by examining their correlation with the maximum acceleration the materials can withstand without spilling from the spoon during transport.
\end{abstract}



\section{Introduction}
\label{sec:introduction}

Granular materials are ubiquitous in robotic manipulation, from food handling and agriculture to construction and manufacturing, yet their complex physical behavior makes them particularly challenging to manipulate reliably. Unlike rigid objects, their macroscopic response emerges from interactions among many particles and depends on properties such as particle size, shape, packing, and compaction~\cite{clement1999rheology}. Reliable perception of these properties is therefore important for manipulation and transportation, where the same action can produce substantially different outcomes across materials.

Existing approaches to granular-material manipulation have primarily relied on either real-world sensory feedback or physics-based simulation. Real-world methods have used visual, tactile, and vibrational signals to estimate quantities such as scooped mass and material flow~\cite{clarke2018learning,takahashi2025scu}. While these approaches demonstrate that interaction signals contain rich information about granular materials, they often target task-specific quantities or predefined material categories rather than generalizable physical properties. Simulation-based approaches provide an alternative for modeling granular-material interactions without extensive real-world data~\cite{wei2025physics,tuomainen2022manipulation}; however, accurately capturing granular behavior remains challenging because of the high-dimensional particle state and complex inter-particle interactions. Moreover, transferring simulation-based models to real materials requires correspondence between simulated and physical material behavior, often involving material-specific calibration. These limitations motivate learning physical representations directly from real interaction data.

\begin{figure}[!t]
    \centering
    \includegraphics[width=0.9\columnwidth]{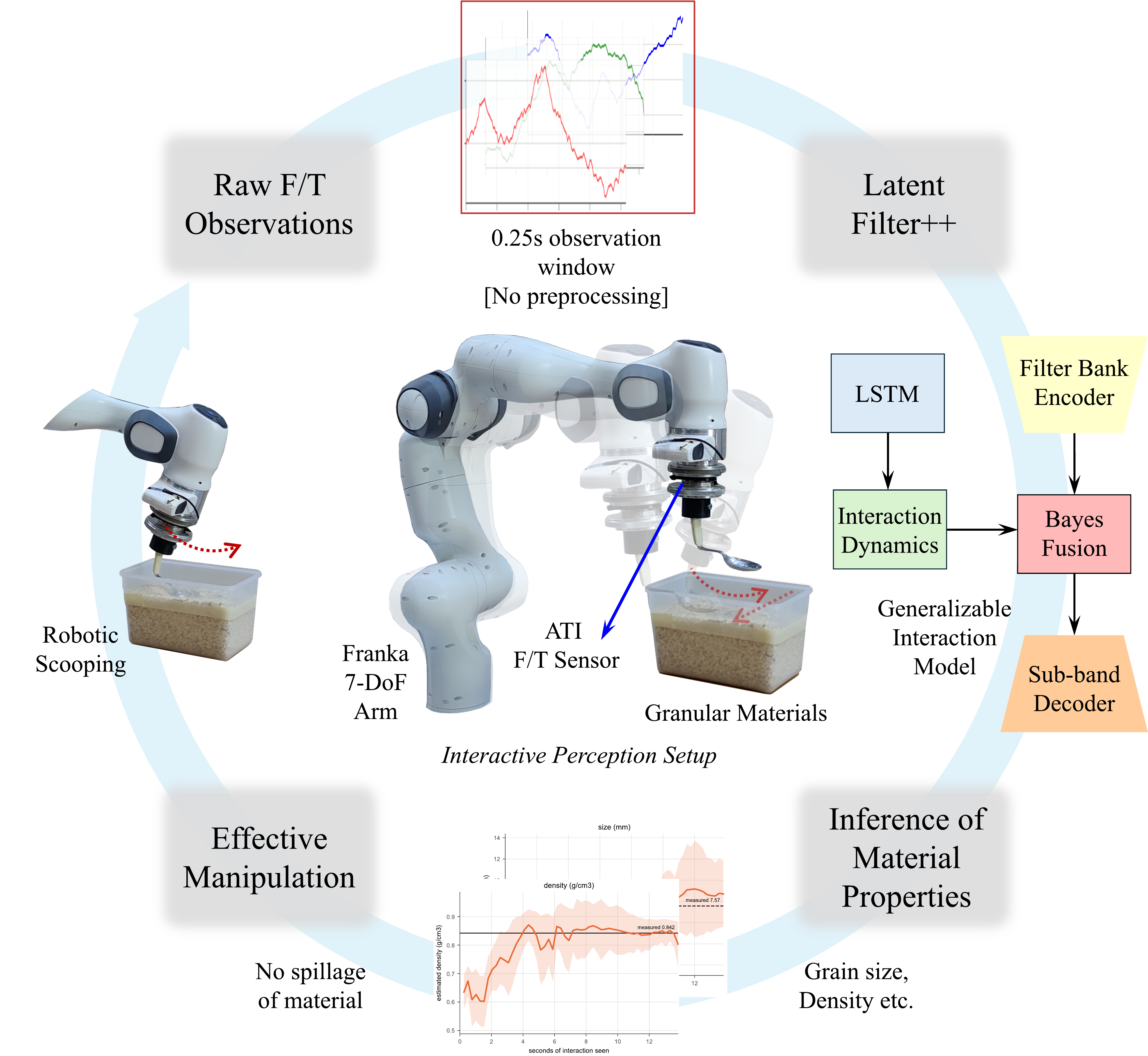}
    \caption{Problem setup for interactive perception of granular materials.}
    \label{fig:problem_setup}
\end{figure}

A key challenge is to extract such properties from the temporal evolution of the interaction rather than from isolated observations, i.e., through interactive perception. In this work, we consider scooping with a spoon as the interaction. When interaction forces during scooping are measured with a force-torque (F/T) sensor, the resulting F/T signal combines slowly varying loads arising from bulk interaction with higher-frequency fluctuations caused by grain-scale contacts. Humans similarly infer material properties during interaction, while different populations of mechanoreceptors exhibit distinct sensitivities to the temporal and frequency characteristics of skin deformation~\cite{johansson2009coding,weber2013spatial}. Inspired by this organization, we introduce a learnable frequency--temporal filter-bank encoder that explicitly determines \emph{which frequencies} to emphasize and \emph{over what time intervals} to integrate. This provides a structured inductive bias for learning from limited interaction data while avoiding handcrafted signal decomposition or dataset-specific preprocessing.

Building on this encoding technique, we extend the latent-filter framework of Dutta et al.~\cite{dutta2026texture} into a causal sequential model for granular-material perception. The proposed \textit{Latent Filter++} operates directly on raw F/T measurements and maintains a temporally evolving representation of the interaction and material properties. This is particularly important for granular materials, for which some physical properties can be measured independently, while others are difficult to characterize directly and may only become apparent through interaction. Rather than requiring explicit supervision for every relevant property, the model learns a latent representation from the observed interaction dynamics. We introduce an innovation pathway and observation masking to encourage the latent representation to capture temporal interaction dynamics, and explicitly evaluate its ability to generalize to materials unseen during training. Separating the learned representation from the property-specific readout further allows adaptation to a new material using only a small number of interaction samples, without retraining the interaction model.

This study makes the following contributions:
\begin{enumerate}
    \item We introduce a bio-inspired frequency--temporal filter-bank encoder that provides a structured inductive bias for F/T measurements without hand-crafted preprocessing, with more consistent performance across observation-window lengths than CNN encoders.
    
    \item We extend the sequential latent-filter framework as \textit{Latent Filter++} with an innovation pathway and observation masking to enable causal online inference from raw F/T measurements, improve latent representation learning, and infer physical properties from interaction dynamics.
    
    \item We demonstrate that the learned representation captures physically meaningful variation, enabling estimation of grain size and bulk density for unseen materials and adaptation of property-specific mappings from a small number of interaction samples.
\end{enumerate}

We evaluate the approach on a real robotic setup with $26$ diverse granular materials and demonstrate its downstream utility by showing how the inferred properties are related to maximum accelerations that the materials can withstand without spilling from the spoon during transportation.
\section{Related work}
\label{sec:related_work}

Granular material manipulation has received increasing attention in robotics, spanning tabletop manipulation with robotic arms and field applications with heavy machinery. Tabletop studies have considered scooping and digging~\cite{schenck2017learning, tai2023scone, takahashi2025scu, wei2025physics, wang2026simulation}, pouring~\cite{clarke2018learning, hu2025robocap, tuomainen2022manipulation}, and shaping~\cite{kim2019developing, alatur2023material, xue2023neural, kreis2025interactive}, while field robotics has investigated autonomous excavation~\cite{jud2017planning, zhang2021autonomous, egli2022soil, osa2023learning, egli2024reinforcement} and loading~\cite{dobson2017admittance, dadhich2019field, fernando2019iterative, backman2021continuous, aoshima2021simulation}. Unlike fluids, for which continuum models such as the Navier--Stokes equations provide a well-established basis for modeling macroscopic dynamics, granular materials exhibit behavior that depends strongly on particle-scale interactions, material properties, and packing state.

Consequently, existing approaches typically either simulate granular interactions over a limited range of material properties using particle-based models~\cite{matl2020inferring,tuomainen2022manipulation}, or learn task-specific dynamics and manipulation policies from real or demonstrated interactions~\cite{schenck2017learning,yamaguchi2015pouring,shukla2023framework,wei2025physics}. While these approaches can achieve effective manipulation within their training or simulation regimes, the diversity of granular materials and their sensitivity to interaction conditions make generalization to unseen materials and states challenging.

Accurate material perception can provide information for adapting manipulation strategies, but obtaining physical properties from interaction remains challenging. Previous work has used audio and vibration signals to recognize materials or estimate task-relevant quantities. Audio recordings have been used to identify materials inside closed containers after shaking or rotation~\cite{eppe2018deep, jin2019open}, while vibration measurements have been used for material recognition and characterization~\cite{chen2016learning, wang2023stev}. GelSight-based vibration sensing has further been used to estimate properties such as particle size and shape~\cite{guo2023estimating}. Other approaches infer material properties indirectly from controlled pouring and the resulting material spread observed by vision~\cite{matl2020inferring}. More closely related to our setting, force--torque sensing has been used to characterize granular materials during physical interaction~\cite{zhang2024understanding, zhang2025joint, hynninen2025interactive}. In particular, Hynninen et al.~\cite{hynninen2025interactive} investigated frequency-dependent patterns in F/T signals for material identification, motivating our use of an explicit frequency-domain representation. However, existing approaches primarily focus on material recognition or estimation of individual properties from predefined observation windows. Such window-based processing is not directly suited to interactive perception, where observations arrive sequentially and material properties must be inferred online as the interaction unfolds. This calls for a representation that can accumulate information causally over time rather than relying on a fixed observation sequence.

Sequential models provide a natural framework for this setting. Autoregressive models~\cite{touchandgo2022} have extended tactile perception to temporal settings, but primarily formulate the problem as next-step prediction without explicitly representing the underlying interaction state or action effects. Deep state-space models (DSSMs) with variational inference provide a complementary framework for learning latent representations from sequential observations while explicitly modeling latent dynamics~\cite{latentmatters, krishnan2015deep}. More recently, Dutta et al.~\cite{dutta2026texture, dutta2026embodied} introduced a latent-filter formulation for inferring physical properties from interaction signals. However, their approach relies on heuristic preprocessing to extract spatial and frequency components from the input signals, introducing non-causal dependencies that limit direct online deployment. Moreover, its ability to generalize to materials outside the training distribution was not explicitly evaluated.

In this work, we build on the latent-filter framework to develop a causal sequential representation for interactive perception of granular materials. The proposed approach operates directly on raw F/T measurements through a learnable frequency--temporal filter-bank representation and maintains a temporally evolving latent representation as observations are acquired during scooping. We explicitly evaluate generalization to previously unseen materials and few-shot adaptation of property-specific readouts. Finally, we connect material perception to downstream manipulation by examining the relationship between the inferred properties and acceleration limits for spill-free material transport.

\section{Proposed Method}
\label{sec:method}
\begin{figure*}[t!]
    \centering
    \includegraphics[width=0.8\textwidth]{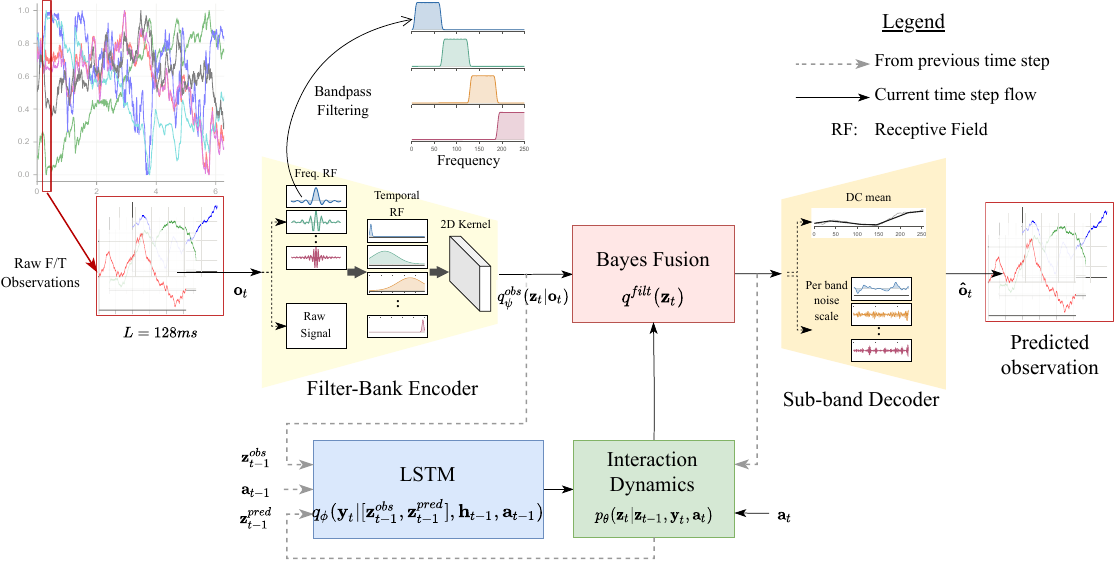}
    \caption{Proposed Latent Filter++ architecture for unsupervised deep state-space learning and inference.}
    \label{fig:framework}
\end{figure*}

\subsection{Problem Definition}
\label{subsec:probdef}
We consider interactive perception of granular materials, where a 7-DoF robotic manipulator equipped with a spoon scoops granular material from a box. We infer the material properties using only kinesthetic sensing from a 6-axis Force/Torque (F/T) sensor mounted between the end-effector and spoon. The robot--material interaction is modeled as a discrete-time nonlinear dynamical system with kinesthetic observations $\mathbf{o}_{1:T} = (\mathbf{o}_1, \mathbf{o}_2, .., \mathbf{o}_T), \mathbf{o}_t \in \mathcal{O} \subset \mathbb{R}^{n_o}$ and actions $\mathbf{a}_{1:T} = (\mathbf{a}_1, \mathbf{a}_2, .., \mathbf{a}_T), \mathbf{a}_t \in \mathcal{A} \subset \mathbb{R}^{n_a}$.

\subsection{Deep State-Space Modeling: Latent Filter++}
Building upon prior work, we model the robot--material interaction using a deep state-space model (DSSM)~\cite{karl2017deep}, with low-dimensional latent states $\mathbf{s}_{1:T}=(\mathbf{s}_1,\ldots,\mathbf{s}_T)$, where $\mathbf{s}_t\in\mathbb{R}^{n_s}$. The model describes the joint distribution $p(\mathbf{o}_{1:T},\mathbf{s}_{1:T}\mid\mathbf{a}_{1:T})$, with inference performed through amortized variational approximation of the latent posterior.

Inferring latent mechanical properties from short interaction sequences is challenging because they are only indirectly observable~\cite{control1}. The F/T sensor measures the aggregate effect of multiple granular contacts, making it difficult to isolate material properties from bulk interaction forces. We therefore build upon the Latent Filter~\cite{dutta2026texture}, which hierarchically structures the latent space as $\mathbf{s}_t=(\mathbf{z}_t,\mathbf{y}_t)$, with $n_s=n_z+n_y$ and where $\mathbf{z}_t\in\mathbb{R}^{n_z}$ denotes a directly observable state, recoverable from a single observation, while $\mathbf{y}_t\in\mathbb{R}^{n_y}$ captures indirectly observable physical properties requiring temporal evidence. We assume that $\mathbf{z}_t$ generates the current time-varying observations, whereas $\mathbf{y}_t$ remains constant within a scoop. Under the Markov assumption, the resulting generative model is
\begin{align}
\label{eq:generativemodelprop}
p(\mathbf{o}_{1:T}|\mathbf{a}_{1:T}) = \iint \prod_{t=2}^T &
p(\mathbf{o}_t|\mathbf{z}_{t}) \,
p(\mathbf{z}_t|\mathbf{z}_{t-1},\mathbf{y}_t,\mathbf{a}_t) \nonumber \\
& \times p(\mathbf{y}_t|\mathbf{y}_{t-1},\mathbf{a}_{t-1})
\,d\mathbf{y}_{t-1}d\mathbf{z}_{t-1}
\end{align}

We adopt the corresponding amortized variational inference framework,
\begin{align}
q(\mathbf{z}_{1:T},\mathbf{y}_{1:T}|\mathbf{o}_{1:T},\mathbf{a}_{1:T})
\approx
p(\mathbf{z}_{1:T},\mathbf{y}_{1:T}|\mathbf{o}_{1:T},\mathbf{a}_{1:T})
\end{align}
and incorporate the filtering structure through
\begin{align}
\label{eq:qfilter}
&q_{\theta}(\mathbf{z}_{1:T},\mathbf{y}_{1:T}|\mathbf{o}_{1:T},\mathbf{a}_{1:T})
=
q_{\psi}(\mathbf{z}_1|\mathbf{o}_1)q_{\phi}(\mathbf{y}_1)
\prod_{t=2}^{T}q(\mathbf{z}_t,\mathbf{y}_t) \\
&q(\mathbf{z}_t,\mathbf{y}_t)
\propto
\underbrace{q_{\psi}(\mathbf{z}_t|\mathbf{o}_t)
p_{\theta}(\mathbf{z}_t|\mathbf{z}_{t-1},\mathbf{y}_t,\mathbf{a}_t)}_{q^{filt}(\mathbf{z}_t)}
q_{\phi}(\mathbf{y}_t|\mathbf{z}_{t-1},\mathbf{y}_{t-1},\mathbf{a}_{t-1}) \nonumber
\end{align}
Here, $q_{\psi}(\mathbf{z}_t|\mathbf{o}_t)$ provides an amortized inverse measurement model, while $p_{\theta}(\mathbf{z}_t|\mathbf{z}_{t-1},\mathbf{y}_t,\mathbf{a}_t)$ captures predictive interaction dynamics. The distribution $q_{\phi}(\mathbf{y}_t|\mathbf{z}_{t-1},\mathbf{y}_{t-1},\mathbf{a}_{t-1})$ propagates belief over the slowly varying physical properties. Thus, $q^{filt}(\mathbf{z}_t)$ combines instantaneous observations with predictive dynamics to provide a filtered estimate of the directly observable state.

The ELBO and its derivation closely follow~\cite{dutta2026texture}:
\begin{align}
    \label{eq:finalobj}
    \mathcal{F}&(\theta,\phi,\psi) = \;  \mathbb{E}_{q}\Big[ \textstyle\sum_{t=1}^{T} \log p(\mathbf{o}_t|\mathbf{z}_t) \Big] \nonumber \\
    & - \lambda \textstyle\sum_{t=2}^{T} \textsc{kl}\big[ q^{\text{filt}}(\mathbf{z}_t) \,\|\, p_{\theta}(\mathbf{z}_{t}|\mathbf{z}_{t-1},\mathbf{y}_{t},\mathbf{a}_{t}) \big] \nonumber \\
    & - \beta \textstyle\sum_{t=2}^{T} \textsc{kl}\big[ q_\phi(\mathbf{y}_t|\cdot) \,\|\, p(\mathbf{y}_t|\mathbb{N}) \big]
\end{align}
The coefficient $\beta$ is annealed from $0.05$ to $1$, while $\lambda=1$. The prior $p(\mathbf{y}_t|\mathbb{N})$ is a hierarchical conditional prior learned from object labels~\cite{dutta2026texture}. It regularizes the latent representation without imposing a fixed class structure, allowing the latent space to remain continuous and capture variation within and across categories. Our approach introduces three modifications to the prior work \cite{dutta2026texture} in terms of architecture and training procedure and which we refer to as Latent Filter++ (see Fig.~\ref{fig:framework}) to improve representation stability and enable learning directly from raw F/T observations (see Sec.~\ref{sec:exp}).

\subsubsection{Innovation pathway}
In prior work, $q_{\phi}(\mathbf{y}_t|\cdot)$ is parameterized by an LSTM that receives the sampled filtered state $\mathbf{z}_{t-1} \sim q^{filt}(\mathbf{z}_{t-1})$. However, Eq.~\ref{eq:qfilter} combines the measurement and transition distributions, discarding their disagreement. This discrepancy, captured by the innovation
\begin{align}
\label{eq:innovation}
&\boldsymbol{\nu}_{t-1}
=
\mathbf{z}_{t-1}^{\mathrm{obs}}
-
\mathbf{z}_{t-1}^{\mathrm{pred}}, \\
\mathbf{z}_{t-1}^{\mathrm{obs}}
\sim q_{\psi}(\mathbf{z}_{t-1}&|\mathbf{o}_{t-1}), \;
\mathbf{z}_{t-1}^{\mathrm{pred}}
\sim
p_{\theta}(\mathbf{z}_{t-1}|
\mathbf{z}_{t-2},\mathbf{y}_{t-1},\mathbf{a}_{t-1})
\end{align}
quantifies the discrepancy between predicted and observed interaction responses and can provide information about material properties. We therefore provide the LSTM with both samples and the action $\left[\mathbf{z}_{t-1}^{\mathrm{obs}}, \mathbf{z}_{t-1}^{\mathrm{pred}}, \mathbf{a}_{t-1}\right]$, increasing its input dimension from $n_z+n_a$ to $2n_z+n_a$. These quantities are taken from step $t-1$ to preserve causality. At the initial step, the standard normal prior is used in place of the unavailable prediction.

\subsubsection{Observation masking}
We introduce observation masking, inspired by masked autoencoders~\cite{he2022masked}, to encourage the model to exploit learned dynamics rather than reconstruct each observation independently. Without masking, each observation can be reconstructed from its corresponding latent encoding, leaving $\mathbf{y}$ weakly constrained since it is not directly used by the decoder and only conditions the transition. We therefore mask the measurement at a fraction $\rho$ of time steps, sampled independently as $\mathrm{Bernoulli}(\rho)$ for each recording and step. The first two observations remain unmasked to initialize $\mathbf{y}$ from actual measurements, while masked observations are still decoded and evaluated against the corresponding ground truth.

At a masked step, the model must propagate the latent state through $p_{\theta}(\mathbf{z}_t|\mathbf{z}_{t-1},\mathbf{y}_t,\mathbf{a}_t)$, encouraging $\mathbf{y}_t$ to capture material information relevant to predicting interaction dynamics. The masked measurement is represented by a sample from the standard normal prior, $\mathbf{z}_t^{\mathrm{obs}} \sim \mathcal{N}(\mathbf{0},\mathbf{I})$, which is fused with the predictive distribution according to Eq.~\ref{eq:qfilter}. This preserves uncertainty at the masked step rather than treating the prediction as observed.

The same prior sample is used in the innovation pathway at the next step. Using the transition prediction instead would give $\boldsymbol{\nu}_{t-1}=\mathbf{0}$, incorrectly indicating perfect agreement rather than missing evidence. Masking is applied only during training; test-time inference uses unmasked observations.

\subsubsection{Bio-inspired mechanosensory filter-bank encoder}
\label{sec:filterbank_encoder}

F/T measurements contain information over coupled frequencies and timescales~\cite{hynninen2025interactive}. Slowly varying components describe sustained loading and quasi-static interaction, while faster components can reflect transient dynamics. This resembles biological mechanosensation, where slowly adapting (SA) afferents respond to sustained deformation and rapidly adapting (RA) afferents to changes and vibration~\cite{johansson2009coding}. Inspired by this organization, we structure the encoder receptive fields along two dimensions: \emph{frequency}, which determines what spectral content is attended to, and \emph{time}, which determines over what interval responses are integrated.

Let $\mathbf{o}_t\in\mathbb{R}^{C\times L}$ denote the observation, where $C=6$ is the number of F/T axes and $L$ is the short observation-window length sampled at $f_s$. The encoder maps the observation to a feature representation, which is subsequently mapped to the parameters of $q_\psi(\mathbf{z}_t|\mathbf{o}_t)$.

A conventional 1D CNN learns patterns within a local temporal neighborhood. We instead make two aspects of this receptive field explicit: a learnable filter bank determines \emph{which frequencies} are attended to, while temporal integration windows determine \emph{over what time interval} responses are aggregated. A lightweight convolutional stage then learns \emph{what combinations} of these frequency--temporal responses are informative for the latent state.

\paragraph{Frequency receptive fields}
Following SincNet~\cite{ravanelli2018sincnet}, each filter is parameterized by its lower and upper cutoff frequencies rather than independently learning all filter coefficients. For the $j$-th filter on axis $c$,
\begin{equation}
g_{c,j}[n]
=
\frac{w[n]}{2B_{c,j}}
\left[
2f^{\mathrm{hi}}_{c,j}
\operatorname{sinc}
\left(
\frac{2f^{\mathrm{hi}}_{c,j}n}{f_s}
\right)
-
2f^{\mathrm{lo}}_{c,j}
\operatorname{sinc}
\left(
\frac{2f^{\mathrm{lo}}_{c,j}n}{f_s}
\right)
\right]
\end{equation}
where $w[n]$ is a Hamming window and $B_{c,j}=f^{\mathrm{hi}}_{c,j}-f^{\mathrm{lo}}_{c,j}$. The cutoffs are learned independently for each F/T axis, yielding $J$ frequency channels per axis. Filters are initialized to tile the spectrum uniformly and no rectification is applied, retaining signed responses. Each filter therefore defines an explicit, learnable frequency receptive field, yielding $CJ$ band signals of length $L$.

\paragraph{Temporal receptive fields}
The temporal organization of an interaction is not known a priori, as phases such as approach, penetration, drag, and release need not occur at fixed positions. We therefore learn where and over what extent to integrate the responses. For discrete time $\tau\in\{0,\ldots,L-1\}$, the $k$-th of $K$ temporal receptive fields is
\begin{equation}
\omega_k(\tau)
=
\frac{
\exp\left[-(\tau-\mu_k)^2/(2\sigma_k^2)\right]
}{
\sum_{\tau'=0}^{L-1}
\exp\left[-(\tau'-\mu_k)^2/(2\sigma_k^2)\right]
},
\end{equation}
where $\mu_k$ and $\sigma_k$ control its location and extent. Narrow fields emphasize localized changes, whereas broader fields integrate sustained interaction. The integration reduces $L$ samples to $K$ temporally localized responses, providing a fixed-size representation for the measurement model.

\begin{figure*}[!ht]
    \centering
    \includegraphics[width=0.8\textwidth]{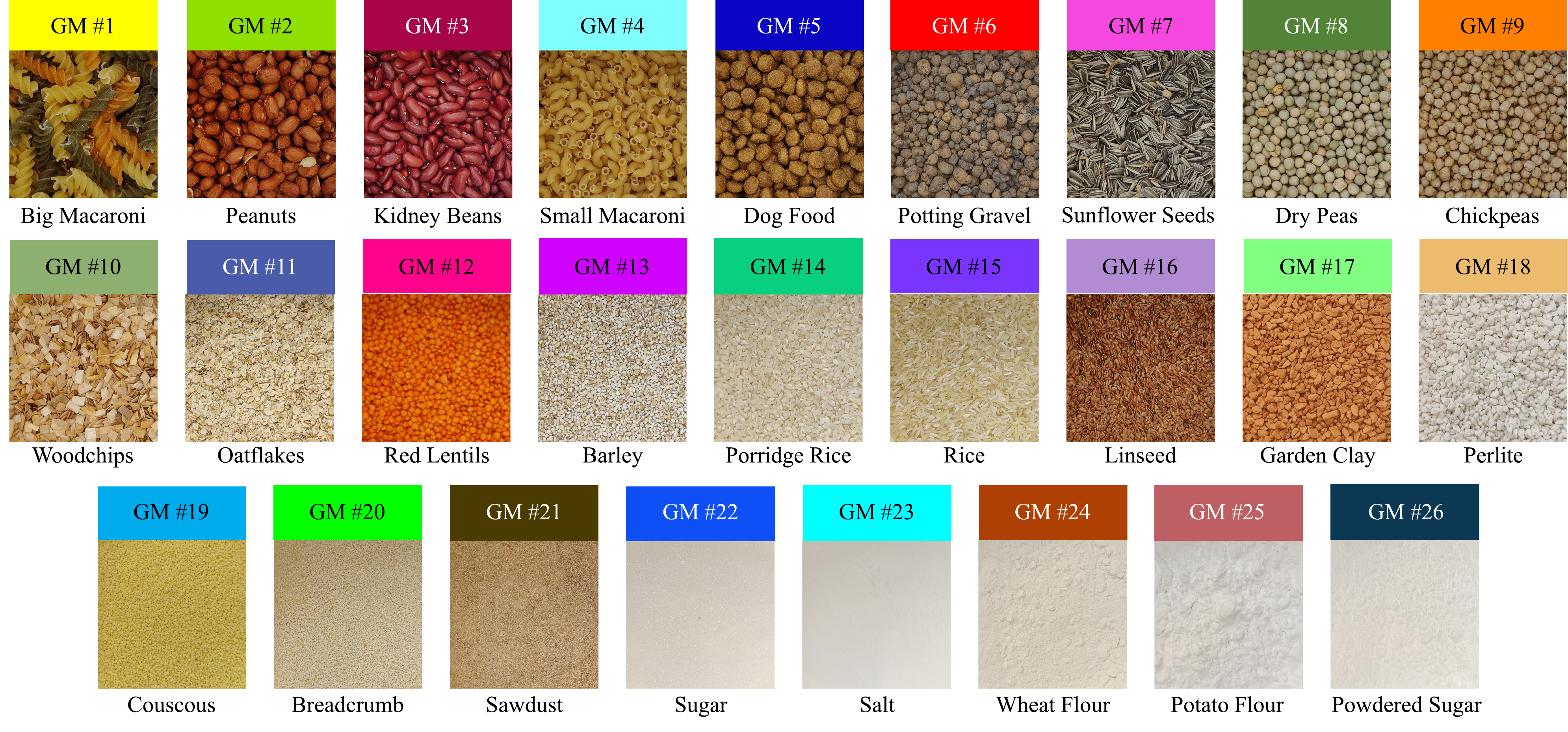}
    \caption{Selected granular materials (GM) used in the experiments. The color labels correspond to the latent-feature UMAP plots.}
    \label{fig:granular_material_samples}
\end{figure*}

\paragraph{Assembling the representation}
The frequency and temporal receptive fields are realized through two parallel pathways. The first operates directly on the $C$ raw F/T axes, preserving sustained and low-frequency components independently of the learned spectral decomposition. The second operates on the $CJ$ band signals. In both pathways, a lightweight grouped convolution produces $W$ local responses, which are integrated by the same learned temporal receptive fields. Each axis or frequency band is represented by a $K\times W$ table, and the pathways are stacked to form a tensor with $C+CJ$ channels. Frequency is represented through channels rather than an additional spatial dimension.

A two-dimensional convolution over the resulting $K\times W$ representation mixes these channels and local temporal responses. This is the first stage where the two pathways and frequency bands interact, while preceding stages retain the explicit receptive-field structure. A linear head then maps the representation to the parameters of $q_\psi(\mathbf{z}_t|\mathbf{o}_t)$.

The resulting encoder combines an explicit receptive-field structure with learned feature composition. Each frequency and temporal field requires two parameters, giving $2CJ+2K$ receptive-field parameters, while the subsequent convolutions learn task-relevant combinations. The prescribed filter length also imposes a lower bound on the observation-window length $L$. We evaluate the structured encoder against generic CNNs in Sec.~\ref{sec:exp}.

\subsubsection{Sub-band decoder}
The observation model \(p(\mathbf{o}_t|\mathbf{z}_t)\) mirrors the frequency--temporal structure of the encoder. Since temporal integration removes phase information, the phase-resolved vibration waveform cannot be recovered from \(\mathbf{z}_t\) and is therefore not explicitly reconstructed. Instead, the decoder predicts a slowly varying mean \(\bar{o}_c(\mathbf{z}_t)\) for each F/T axis and one energy \(E_{c,b}(\mathbf{z}_t)\) for each frequency band:
\begin{equation}
\mathbf{o}_{t,c} \sim \mathcal{N}\!\left(\bar{o}_c(\mathbf{z}_t),\boldsymbol{\Sigma}_c(\mathbf{z}_t)\right), \;
S_c(f)=\sum_{b=1}^{J}E_{c,b}(\mathbf{z}_t)^2\left|H_{c,b}(f)\right|^2
\end{equation}
Here, \(\boldsymbol{\Sigma}_c(\mathbf{z}_t)\in\mathbb{R}^{L\times L}\) is a Toeplitz covariance matrix whose autocorrelation function is obtained by the inverse Fourier transform of the power spectral density \(S_c(f)\) \cite{brockwell1991time}. This allows uncertainty to be concentrated in particular frequency ranges rather than represented by a single noise scale. The decoder-side filter bank \(H_{c,b}(f)\) is parameterized independently of the encoder bank, preventing the reconstruction objective from imposing a particular frequency partition on the encoder; instead, the encoder bands are shaped by frequency components informative for the latent state.
\section{Experiments \& Results}
\label{sec:exp}

\subsection{Robotic Setup, Data Collection \& Training of \textit{Latent Filter}}
We evaluate the proposed approach on a curated set of $26$ everyday granular materials, shown in Fig.~\ref{fig:granular_material_samples}. The materials span five morphological classes, including coarse to fine: hard pieces (macaroni, peanuts, kidney beans), pebbles and seeds (potting gravel, sunflower seeds, chickpeas), small grains and flakes (barley, rice, oat flakes), fine granular materials (couscous, breadcrumbs, sawdust), and powders (salt, sugar, wheat flour). A Franka Emika Panda carries a spoon, with a 6-axis ATI F/T sensor mounted between the end-effector and spoon, as shown in Fig.~\ref{fig:problem_setup}.

Three scooping demonstrations were provided kinesthetically, and the recorded trajectories were filtered and reduced to fixed waypoints. Each interaction starts from a fixed height, with scooping initiated upon contact detection. We collected $40$--$50$ interactions per material while varying the initial position, resulting in $1316$ recordings. Each recording contains $14$\,s of six-axis F/T measurements sampled at $f_s=500$\,Hz. These are presented to the filter as $T=54$ observations of $L=128$ samples ($256$\,ms), such that at step $t$ the model has access only to past and current measurements. For each material, we measured bulk density, mean grain size (approximated for powders), and angle of repose using laboratory measurements.

\subsubsection{Analysis of the filter-bank encoder}
Before evaluating the complete latent filter, we isolate the encoder to assess whether explicitly structured frequency and temporal receptive fields provide a useful representation of the F/T signal. We use supervised material recognition as a controlled comparison with existing spectrum-based methods. All encoders are trained under the same procedure and followed by a single linear readout. Each model is trained across three seeds, determining both parameter initialization and the train/validation/test partition.

We compare three alternatives. Baseline~1 is GmClass~\cite{zhang2025joint}, which combines a spectrum encoder with a language-model embedding of the material name. Baseline~2 is a generic two-layer CNN. Baseline~3 is a two-layer adaptation of SincNet~\cite{ravanelli2018sincnet}, using a single filter bank shared across axes and without explicit temporal receptive fields. We vary the observation window from $1$ to $50\%$ of the recording ($0.14$--$7$\,s) to evaluate the effect of available interaction history on classification accuracy.

\begin{figure}[!htb]
    \centering
    \includegraphics[width=0.7\columnwidth]{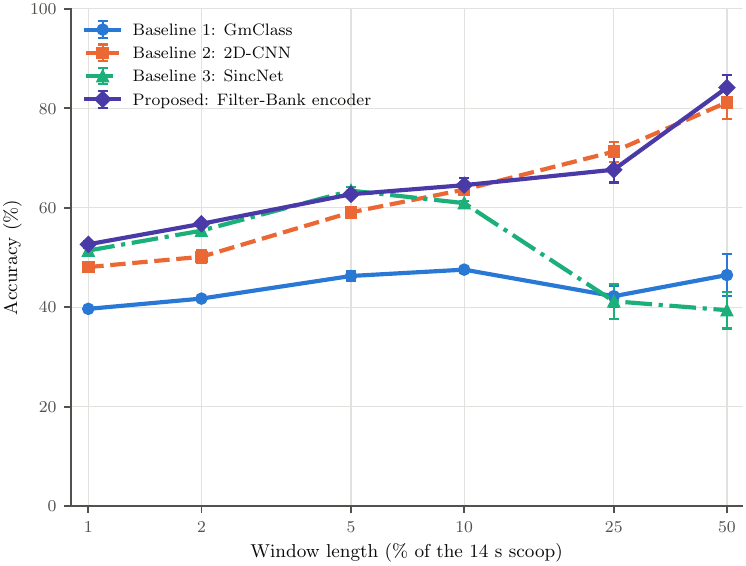}
    \caption{Material classification accuracy as a function of observation-window length for the proposed filter-bank encoder and comparison methods.}
    \label{fig:classification}
\end{figure}

Fig.~\ref{fig:classification} shows that the proposed encoder is more robust to changes in observation-window length. At the shortest window of $0.14$\,s, it achieves $52.6\%$ window accuracy, compared with $48.1\%$ for the two-layer CNN and $39.7\%$ for GmClass. SincNet remains competitive at $5\%$ of the recording, reaching $63.4\%$, but falls to $39.4\%$ at $50\%$. The two-layer CNN improves with window length but starts substantially lower, whereas the proposed encoder improves consistently and reaches $84.2\%$. 

\subsection{Analysis of latent features}
We next evaluate whether the latent representation learned by \textit{Latent Filter++} captures physically meaningful information about the material. We compare it with the prior latent-filter approach~\cite{dutta2026texture}, using the same objective but replacing the proposed filter-bank encoder and decoder with a generic CNN encoder and de-convolutional decoder, and removing observation masking and the innovation pathway. Neither model receives grain size or bulk density as input; the material label enters only through the prior $p(\mathbf{y}_t|\mathbb{N})$. For reference, we also train a supervised model that directly regresses both properties from the same observation buffers using a two-layer CNN encoder followed by an LSTM.

\begin{figure*}[!htb]
    \centering
    \includegraphics[width=0.95\textwidth]{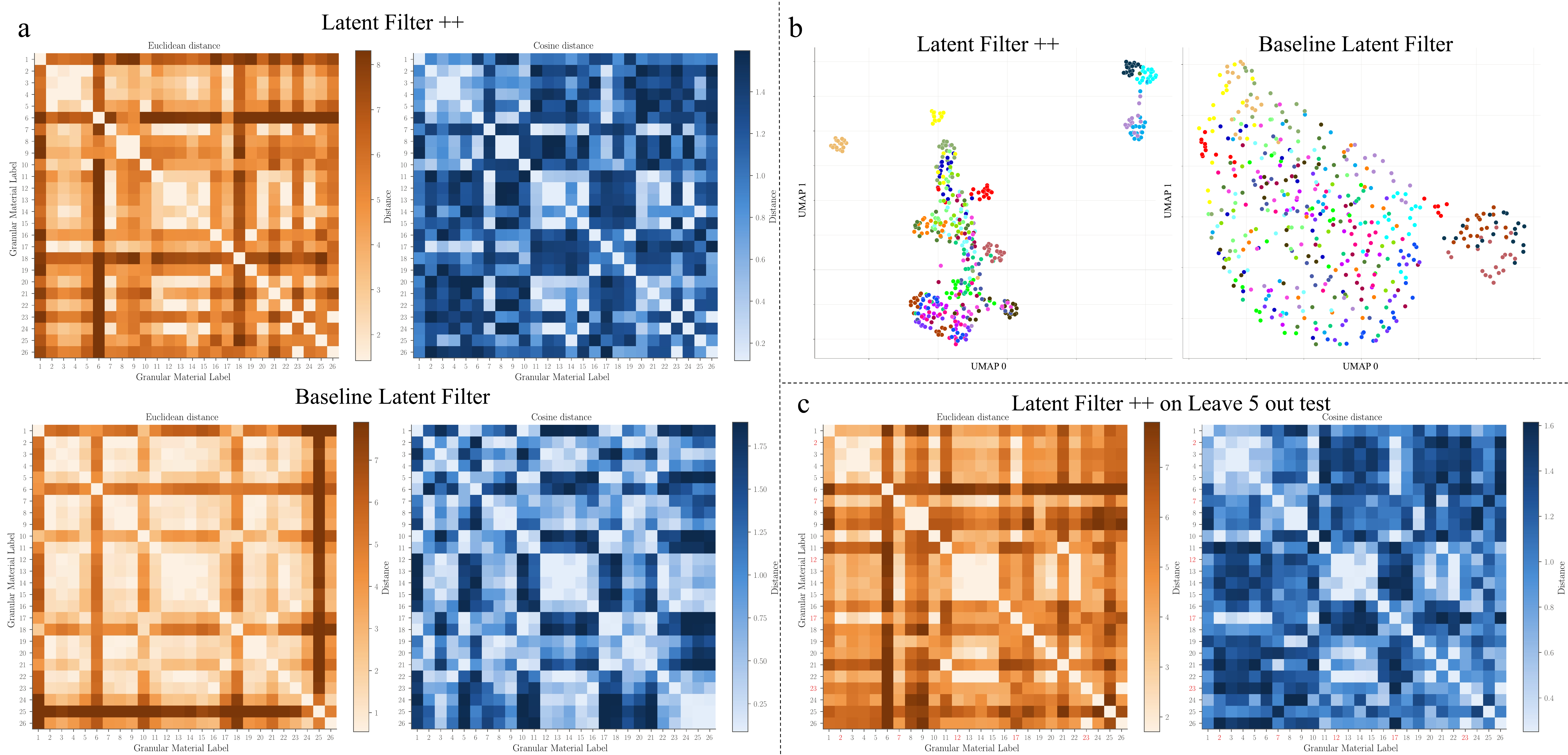}
    \caption{Visualization of the learned latent representation. a) Pairwise latent distances and b) two-dimensional UMAP projections are shown for the evaluated models. c) Distances on leave 5 out test, with left out material label highlighted in \textcolor{red}{red}.}
    \label{fig:exp2latentmap}
\end{figure*}

After training, the DSSM models are frozen and a single kernel ridge regression (KRR) maps $\mathbf{y}$ to each physical property, separating representation learning from the property-specific readout and enabling subsequent online adaptation. Fig.~\ref{fig:exp2latentmap} visualizes the resulting latent spaces using their pairwise distance structures and 2D UMAP projections.

\begin{figure}[!htb]
    \centering
    \includegraphics[width=\columnwidth]{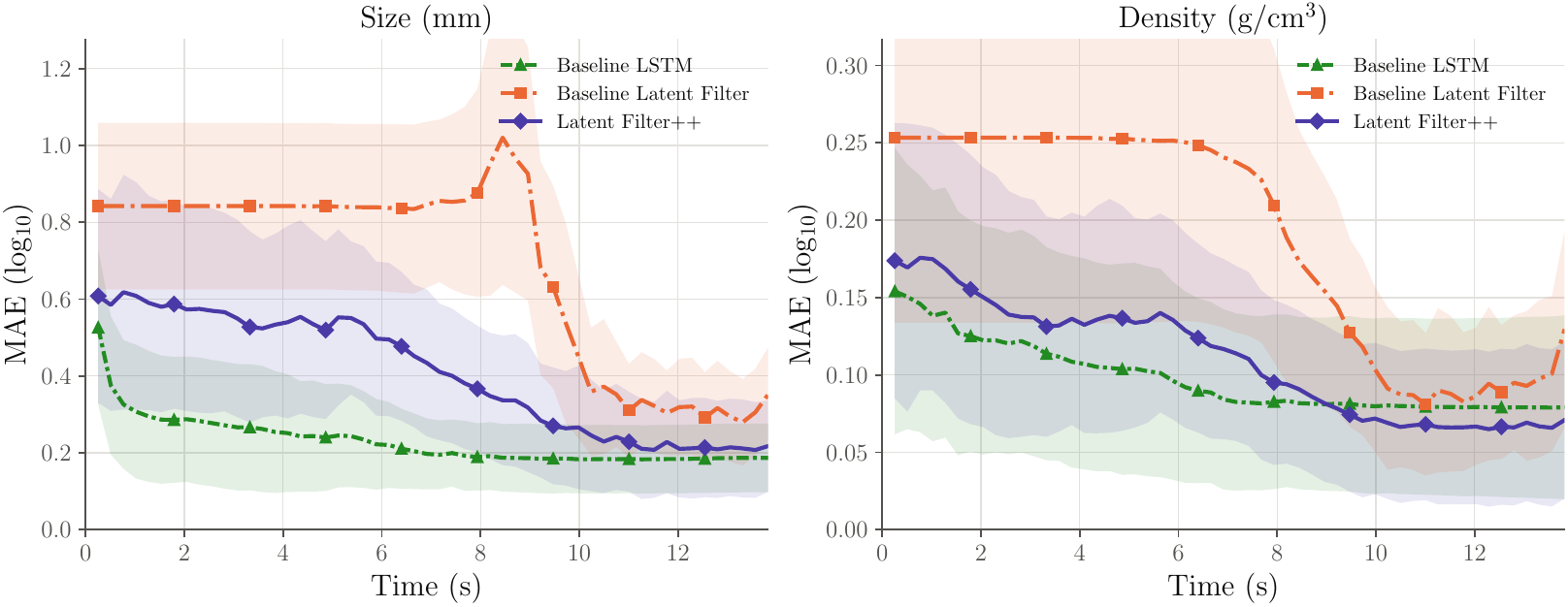}
    \caption{Temporal prediction of grain size and bulk density using \textit{Latent Filter++}, the prior latent-filter approach, and the supervised two-layer CNN+LSTM.}
    \label{fig:exp2temporalprediction}
\end{figure}

The proposed representation captures information relevant to both properties before the end of the scoop (Fig.~\ref{fig:exp2temporalprediction}). At $13.8$\,s, \textit{Latent Filter++} achieves errors of $0.22$ for grain size and $0.07$ for bulk density, compared with $0.35$ and $0.13$ for the prior approach, corresponding to reductions of $38\%$ and $45\%$, respectively. The supervised two-layer CNN+LSTM achieves lower in-distribution grain-size error ($0.19$) and comparable density error ($0.08$). Its performance on unseen materials is evaluated next.

\subsection{Evaluation of generalization performance}
The central test of the learned representation is whether it transfers to materials that were never observed during training. We therefore withhold materials entirely from training and evaluate all $26$ leave-one-out cases, together with five sets each for leave-three-out and leave-five-out. Table~\ref{tab:exp3_mae} summarizes the resulting errors. The supervised model achieves lower grain-size error in distribution, while density errors are comparable. Once materials are withheld, the proposed model performs better for both properties; at leave-five-out, the errors are $0.50$ versus $0.63$ for grain size and $0.17$ versus $0.20$ for density. This indicates that the learned representation transfers better to unseen materials than the directly supervised model.

\begin{figure}[!htb]
    \centering
    \includegraphics[width=\columnwidth]{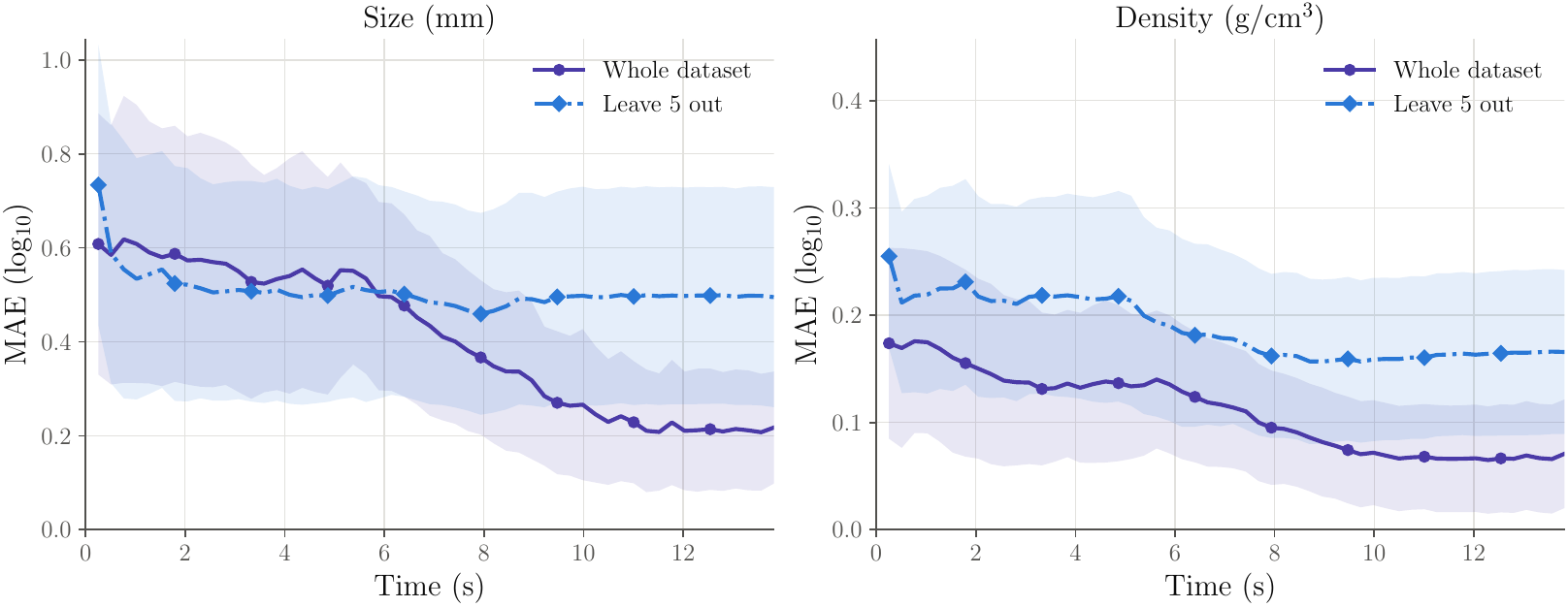}
    \caption{Temporal prediction of latent features for an out-of-distribution leave-five-out test set.}
    \label{fig:exp3generalization}
\end{figure}

We next examine whether the latent representation itself remains consistent as materials are withheld. Table~\ref{tab:exp3_representation} measures agreement between the latent space obtained from each held-out run and that learned from the full dataset, using views invariant to rotation, reflection, scale, and translation. Agreement decreases gradually from $0.78$ at leave-one-out to $0.69$ at leave-five-out, indicating that the latent organization remains largely preserved as increasingly many materials are removed. This is also reflected qualitatively in the distance matrices in Fig.~\ref{fig:exp2latentmap}c) and also in Fig.~\ref{fig:exp3generalization}.

This consistency supports few-shot adaptation to new materials without retraining the latent filter. A small number of scoops can instead anchor the property readout through efficient KRR training. Table~\ref{tab:exp3_anchoring} evaluates this setting. For leave-one-out, grain-size error decreases from $0.46$ with a single anchor to $0.23$ with forty anchors. Even one scoop substantially reduces the gap, indicating that the property-specific readout can be adapted with a small amount of interaction rather than retraining the sequential model.

\begin{table}[t]
  \centering
  \caption{Error on materials held out of training ($\log_{10}$, mean $\pm$ s.d.). Grain size and bulk density are predicted in $\log_{10}$, so an error reads as a multiplicative factor: $0.30$ is a factor of two out. The spread is across held out materials, not across seeds. Lower is better; the better of the two models in each row is bold.}
  \label{tab:exp3_mae}
  \setlength{\tabcolsep}{5pt}
  \footnotesize
  \begin{tabular}{lcccc}
    \toprule
    & \multicolumn{2}{c}{Grain size} & \multicolumn{2}{c}{Bulk density} \\
    \cmidrule(lr){2-3} \cmidrule(lr){4-5}
    Condition & LSTM & Proposed & LSTM & Proposed \\
    \midrule
    Whole dataset (1) & $\mathbf{0.19}{\scriptstyle\,\pm\,0.18}$ & $0.22{\scriptstyle\,\pm\,0.24}$ & $0.08{\scriptstyle\,\pm\,0.12}$ & $\mathbf{0.07}{\scriptstyle\,\pm\,0.10}$ \\
    Leave 1 out (26) & $0.61{\scriptstyle\,\pm\,0.61}$ & $\mathbf{0.54}{\scriptstyle\,\pm\,0.48}$ & $0.19{\scriptstyle\,\pm\,0.17}$ & $\mathbf{0.16}{\scriptstyle\,\pm\,0.16}$ \\
    Leave 3 out (5) & $0.77{\scriptstyle\,\pm\,0.14}$ & $\mathbf{0.61}{\scriptstyle\,\pm\,0.14}$ & $0.14{\scriptstyle\,\pm\,0.06}$ & $\mathbf{0.12}{\scriptstyle\,\pm\,0.03}$ \\
    Leave 5 out (5) & $0.63{\scriptstyle\,\pm\,0.13}$ & $\mathbf{0.50}{\scriptstyle\,\pm\,0.08}$ & $0.20{\scriptstyle\,\pm\,0.07}$ & $\mathbf{0.17}{\scriptstyle\,\pm\,0.06}$ \\
    \bottomrule
  \end{tabular}
\end{table}
\begin{table}[t]
  \centering
  \caption{Agreement between each leave-$N$-out run's latent space and the whole dataset run's, averaged over that condition's cases. $1.0$ is an identical arrangement of the materials and $0.0$ an unrelated one; every view is invariant to rotation, reflection, scale and translation. Higher is better.}
  \label{tab:exp3_representation}
  \begin{tabular}{lcccc}
    \toprule
    Condition & Combined & Euclidean & Cosine & Procrustes \\
    \midrule
    Whole dataset (1) & $0.974$ & $0.979$ & $0.968$ & $0.979$ \\
    Leave 1 out (26) & $0.779$ & $0.819$ & $0.739$ & $0.800$ \\
    Leave 3 out (5) & $0.753$ & $0.801$ & $0.705$ & $0.761$ \\
    Leave 5 out (5) & $0.692$ & $0.751$ & $0.633$ & $0.733$ \\
    \bottomrule
  \end{tabular}
\end{table}

\begin{table*}[t]
  \centering
  \setlength{\tabcolsep}{4pt}
  \small
  \caption{Error on the held out materials after $k$ of their own scoops are added to the readout ($\log_{10}$, mean $\pm$ s.d. across held out materials). The model is never retrained: only the readout sees the anchors. Both properties are predicted in $\log_{10}$, so an error reads as a multiplicative factor.}
  \label{tab:exp3_anchoring}
  \begin{tabular}{lcccccc}
    \toprule
    & \multicolumn{3}{c}{Grain size} & \multicolumn{3}{c}{Bulk density} \\
    \cmidrule(lr){2-4} \cmidrule(lr){5-7}
    Anchors & Leave 1 out (26) & Leave 3 out (5) & Leave 5 out (5) & Leave 1 out (26) & Leave 3 out (5) & Leave 5 out (5) \\
    \midrule
    $k{=}1$  & $0.455{\scriptstyle\,\pm\,0.423}$ & $0.569{\scriptstyle\,\pm\,0.112}$ & $0.427{\scriptstyle\,\pm\,0.052}$ & $0.140{\scriptstyle\,\pm\,0.135}$ & $0.113{\scriptstyle\,\pm\,0.027}$ & $0.141{\scriptstyle\,\pm\,0.045}$ \\
    $k{=}5$  & $0.355{\scriptstyle\,\pm\,0.321}$ & $0.460{\scriptstyle\,\pm\,0.097}$ & $0.329{\scriptstyle\,\pm\,0.016}$ & $0.115{\scriptstyle\,\pm\,0.091}$ & $0.103{\scriptstyle\,\pm\,0.024}$ & $0.110{\scriptstyle\,\pm\,0.031}$ \\
    $k{=}10$ & $0.304{\scriptstyle\,\pm\,0.259}$ & $0.392{\scriptstyle\,\pm\,0.061}$ & $0.286{\scriptstyle\,\pm\,0.032}$ & $0.100{\scriptstyle\,\pm\,0.070}$ & $0.091{\scriptstyle\,\pm\,0.021}$ & $0.101{\scriptstyle\,\pm\,0.034}$ \\
    $k{=}20$ & $0.267{\scriptstyle\,\pm\,0.212}$ & $0.345{\scriptstyle\,\pm\,0.049}$ & $0.247{\scriptstyle\,\pm\,0.046}$ & $0.090{\scriptstyle\,\pm\,0.064}$ & $0.082{\scriptstyle\,\pm\,0.021}$ & $0.091{\scriptstyle\,\pm\,0.033}$ \\
    $k{=}40$ & $0.231{\scriptstyle\,\pm\,0.167}$ & $0.309{\scriptstyle\,\pm\,0.036}$ & $0.224{\scriptstyle\,\pm\,0.052}$ & $0.077{\scriptstyle\,\pm\,0.053}$ & $0.070{\scriptstyle\,\pm\,0.022}$ & $0.082{\scriptstyle\,\pm\,0.030}$ \\
    \bottomrule
  \end{tabular}
\end{table*}

\subsection{Real-world manipulation task: Spill-free transportation}
Finally, we evaluated the importance of the estimated material properties for a real-world manipulation task, namely, spill-free granular material transportation. We measured the maximum acceleration the material can withstand without spilling from the spoon and analyzed correlations between the measured maximum acceleration and grain size and density. We selected a subset of 12 materials from the 26 for this test (red lentils, linseeds, rice, porridge rice, chickpeas, barley, sunflower seeds, garden clay, couscous, dry peas, breadcrumbs, oatflakes). The materials were selected to span a diverse range of properties, with only those that actually spilled within the robot’s kinematic limits included. We observed the strongest correlation for grain size ($r_s = 0.74$), while the correlation with density was more modest ($r_s = 0.21$). These positive correlations show that material perception results can be used to plan efficient yet safe (spill-free) motion trajectories for granular material transportation. 





\section{Discussion \& Conclusion}
\label{sec:discussion}

We presented an interactive perception framework for estimating granular-material properties from F/T measurements during scooping. \textit{Latent Filter++} combines a learnable frequency--temporal representation with a causal latent dynamical model, enabling online estimation, generalization to unseen materials, and few-shot adaptation. We demonstrated its utility for selecting transport accelerations for efficient, spill-free transportation. We further show that bio-inspired signal processing provides an effective inductive bias for F/T encoders when large-scale data collection is expensive.

The experiments reveal differences in property observability. Bulk density is estimated more reliably than grain size, whose predictions exhibit higher variance. We observed that angle of repose is poorly estimated from the scooping interaction, suggesting that this property is not sufficiently observable from the available interaction signals. These results highlight that the observability of a material property depends not only on the sensing modality, but also on the nature of the interaction used to probe the material. Compared with the prior latent-filter approach~\cite{dutta2026texture}, \textit{Latent Filter++} provides more stable generalization while eliminating handcrafted signal preprocessing; compared with supervised regression, its primary advantage is improved performance on held-out materials. 

The approach also has limitations. Although the learned representation generalizes to unseen materials, it does not reliably extrapolate beyond the range represented in training, suggesting that a small number of representative materials from each morphological class may be important for robust generalization. The largest leave-one-out errors occur near representation boundaries, including powdered sugar, large macaroni, perlite, and potting gravel. Moreover, few-shot adaptation currently requires manually selected anchor interactions, as latent prediction variance alone is insufficient to autonomously trigger adaptation. These results motivate future work toward autonomous online adaptation, where uncertainty and active interaction selection determine when and how additional observations should be acquired.

\noindent\textbf{Acknowledgment}: AI was used to paraphrase the human-written text and optimize the human-implemented code-base.


\small{\bibliography{biblio}}
\bibliographystyle{IEEEtran}

\end{document}